\documentclass[letterpaper]{article} 
\usepackage{aaai2026}  
\usepackage{times}  
\usepackage{helvet}  
\usepackage{courier}  
\usepackage[hyphens]{url}  
\usepackage{graphicx} 
\usepackage{natbib}  
\usepackage{caption} 
\usepackage{algorithm}
\usepackage{algorithmic}
\usepackage{graphicx}
\usepackage{amsmath}
\usepackage{multirow}
\usepackage{graphicx}
\usepackage{subcaption}
\usepackage{xcolor}
\usepackage{tabularx} 
\usepackage{siunitx} 
\usepackage{makecell} 
\usepackage{booktabs} 
\usepackage{threeparttable}

\usepackage{newfloat}
\usepackage{listings}
\DeclareCaptionStyle{ruled}{labelfont=normalfont,labelsep=colon,strut=off} 
\floatstyle{ruled}
\newfloat{listing}{tb}{lst}{}
\floatname{listing}{Listing}
\title{Learn to Select: Exploring Label Distribution Divergence for In-Context Demonstration Selection in Text Classification}

\author{
    Ye Jiang \textsuperscript{\rm 1}, Taihang Wang\textsuperscript{\rm 1}, Youzheng Liu\textsuperscript{\rm 1}, Yimin Wang\textsuperscript{\rm 2}\thanks{Corresponding Author}, Yuhan Xia\textsuperscript{\rm 3}, Yunfei Long\textsuperscript{\rm 3}\\
}
\affiliations{
    \textsuperscript{\rm 1} College of Information Science and Technology, Qingdao University of Science and Technology\\
    \textsuperscript{\rm 2} College of Data Science, Qingdao University of Science and Technology\\
    \textsuperscript{\rm 1} School of Electronic Engineering and Computer Science, Queen Mary University of London\\

}

\usepackage{bibentry}

\begin{document}

\maketitle

\begin{abstract}
In-context learning (ICL) for text classification, which uses a few input-label demonstrations to describe a task, has demonstrated impressive performance on large language models (LLMs). However, the selection of in-context demonstrations plays a crucial role and can significantly affect LLMs' performance. Most existing demonstration selection methods primarily focus on semantic similarity between test inputs and demonstrations, often overlooking the importance of label distribution alignment. To address this limitation, we propose a two-stage demonstration selection method, \textbf{TopK} + \textbf{L}abel \textbf{D}istribution \textbf{D}ivergence (\textbf{L2D}), which leverages a fine-tuned BERT-like small language model (SLM) to generate label distributions and calculate their divergence for both test inputs and candidate demonstrations. This enables the selection of demonstrations that are not only semantically similar but also aligned in label distribution with the test input. Extensive experiments across seven text classification benchmarks show that our method consistently outperforms previous demonstration selection strategies. Further analysis reveals a positive correlation between the performance of LLMs and the accuracy of the underlying SLMs used for label distribution estimation.
\end{abstract}



\section{Introduction}

In-context learning (ICL) \cite{brown2020language} is an emergent capability of large language models (LLMs), enabling them to make accurate predictions based on only a few input-output demonstrations provided at inference time \cite{dong2024survey}. Compared to standard zero-shot prompting, ICL has shown superior effectiveness in leveraging in-context demonstrations, and has become a new paradigm for tackling a wide range of text classification tasks, including fake news detection \cite{jiang2024large} and natural language inference \cite{xu2024small}.


However, the classification accuracy of LLMs employing ICL is highly sensitive to the choice and ordering of demonstrations. Prior studies \cite{min2022rethinking, liu2022makes} have shown that even minor changes in the order of examples can lead to substantial variability in model predictions. To address this, retrieval-based methods have been proposed \cite{rubin2022learning}, which aim to select demonstrations that are semantically similar to the test input. These approaches have been shown to consistently outperform random selection strategies.

\begin{figure}[t]
  
  \includegraphics[width=1.03\columnwidth]{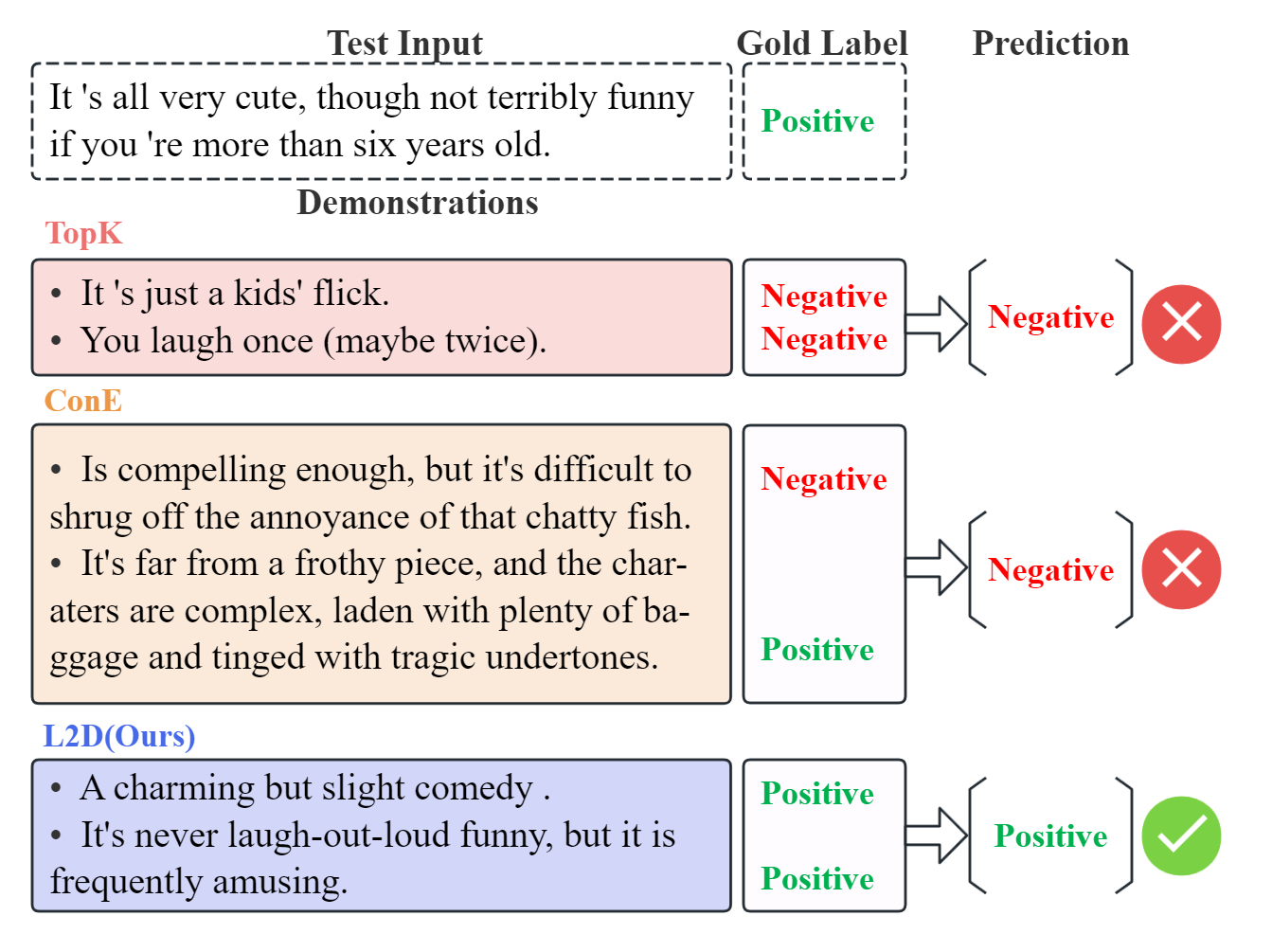}
  \caption{A comparison of 2-shot in-context demonstrations retrieved by different selection methods in SST-2. Although the test input is labeled as having a positive sentiment, the overall semantics are somewhat ambiguous or controversial. Our method effectively captures the adversative conjunction in the demonstrations and aligns the label distributions with that of the test input.
}
  \label{fig:1}
\end{figure}

Furthermore, many studies have examined additional factors that can affect ICL performance in text classification tasks. For example, \citet{iter2023context} suggests that the effectiveness of in-context examples is negatively correlated with the perplexity of a fine-tuned model on the test samples. Similarly, \citet{peng2024revisiting} demonstrates a positive correlation between ICL performance and the model’s comprehension of the test inputs. While these studies have achieved promising results, they predominantly emphasize semantic consistency between the test input and the selected demonstrations. However, semantically similar examples may still contain contradictory or inconsistent labels, undermining their effectiveness. Moreover, test inputs may exhibit label ambiguity due to semantic uncertainty, as shown in Figure~\ref{fig:1}.

A recent study \cite{fei2023mitigating} suggests that the performance improvements observed in LLMs through demonstrations may not primarily arise from accurate input-label pairings. In fact, demonstrations with randomly assigned \cite{min2022rethinking} or symbolic \cite{wei2023symbol} labels have been shown to produce competitive results. This challenges existing selection methods to be more robust in the presence of noisy data, where demonstrations with incorrect labels are often assigned to semantically similar test inputs.

To address the above issues, we propose a two-stage method, denoted as \textbf{TopK} + \textbf{L}abel \textbf{D}istribution \textbf{D}ivergence (\textbf{L2D}). Specifically, we apply the TopK retrieval method \cite{liu2022makes} to extract a candidate pool of demonstrations from the training set, selected based on semantic similarity to the test input. In parallel, we fine-tune a small language model (SLM), such as BERT \cite{devlin2019bert}, on the training data to estimate the label probability distributions of both the candidate demonstrations and the test input. 

Inspired by \citet{guo2021label}, who suggests that labels are often not entirely independent of instances, particularly in the presence of noisy data and label ambiguities, we address this issue by incorporating label distribution alignment into our demonstration selection method. Specifically, we employ Kullback–Leibler (KL) divergence \cite{kullback1951information} to calibrate the discrepancy between the label distributions of the test input and candidate demonstrations. To achieve a more balanced and symmetric alignment, we further apply Jensen–Shannon (JS) divergence \cite{menendez1997jensen}, enabling us to jointly consider both semantic relevance and label distribution consistency. Extensive experiments across multiple LLMs and model scales demonstrate the effectiveness of our method. Additional analysis further reveals a positive correlation between the performance of LLMs and the classification accuracy of the SLMs.

Our main contributions are summarized as follows:

\begin{itemize}
    \item We introduce TopK + L2D,  a two-stage demonstration selection method that integrates semantic similarity and label distribution divergence to \textbf{select in-context demonstrations that are both semantically relevant and label-aligned with the test input}.
    
    \item We investigate the relationship between label distribution consistency and semantic similarity in text. Our findings suggest that \textbf{label distribution helps to disentangle semantic ambiguity, particularly in the presence of noisy data}.
    
    \item We build a lightweight framework to connect the label distribution of demonstration pool with test inputs by simply fine-tuning SLMs. Our findings indicate that \textbf{the classification performance of LLMs is positively correlated with the predictive accuracy of SLMs}.

    \item We achieve state-of-the-art performance across seven classification benchmark tasks. Comprehensive evaluations demonstrate that \textbf{our method consistently outperforms existing baselines across various LLMs and model scales}.
\end{itemize}

\section{Related Work}

LLMs have demonstrated exceptional performance across a wide range of tasks \cite{gao2025leveraging, cai2025improving}. For example, \citet{zhang2025intention} explores the zero-shot capabilities of LLMs and analyzes the impact of different prompting strategies. Similarly, \citet{ziems2024can} outlines a roadmap for employing LLMs as general-purpose tools, offering best practices for prompt engineering and providing a comprehensive evaluation of zero-shot performance. Recent studies \cite{wang2024oop, wang2023label} further reveal that ICL can effectively harness the capabilities of LLMs, leading to enhanced performance across various applications.

However, the performance of ICL has been shown to be unstable \cite{lu2022fantastically, min2022rethinking}, with minor changes in the demonstrations or their ordering often resulting in significant performance fluctuations. To mitigate this issue, a wide range of demonstration selection strategies has been proposed. For example, \citet{liu2022makes} suggests that selecting demonstrations that are semantically similar to each test input can lead to improved ICL performance. \citet{hongjin2022selective} introduces a framework that combines selective annotation with vote-k strategies to enhance the diversity of the demonstration pool. Additionally, \citet{wu2023self} adopts the Minimum Description Length (MDL) principle, using the inference model itself to identify optimal in-context examples.

Meanwhile, other methods have explored additional factors that may influence demonstration selection. For example, \citet{min2022rethinking} finds that LLMs do not heavily rely on correct input-label pairs in demonstrations; even randomly assigned labels can yield relatively strong classification performance. \citet{wei2023symbol} further suggests that ICL performance can be enhanced by replacing natural language labels with arbitrary symbols. \citet{wu2023self} proposes selecting demonstrations that enable lossless compression of test labels. \citet{iter2023context} identifies in-domain demonstrations by computing the cross-entropy difference of test labels using a small model fine-tuned on the demonstrations. Additionally, \citet{peng2024revisiting} observes that demonstration selection is both data and model dependent, and introduces the TopK + ConE method to improve the LLM's understanding of the test input. However, while these approaches focus primarily on the correlation between test inputs and demonstrations, they largely overlook the role of label distribution, a factor that may significantly affect ICL performance, especially in noisy or ambiguous labeling scenarios.

\section{Preliminaries}

We introduce a two‑stage demonstration selection framework, comprising TopK followed by L2D.

In the \textbf{Semantic Retrieval stage}, we employ the TopK algorithm to retrieve a pool of $K$ demonstrations, $D_{pool}^{i} = [d_1, d_2, \ldots, d_K]$, that exhibit the highest semantic similarity to the $i$th test instance $T_{test}^i \in [t_{test}^1, t_{test}^2, \ldots, t_{test}^n]$, where $n$ is the total number of test samples.

In the \textbf{Label Divergence Estimation stage}, a SLM is fine‑tuned on the training set to predict label probability distributions for each test instance, $P_{test}^i \in [p_{test}^1, p_{test}^2, \ldots, p_{test}^n]$,  and for its corresponding demonstration set, $P_{pool}^{i} = [p_1, p_2, \ldots, p_K]$, where demonstrations are initially ranked by their similarity scores.


To address discrepancies between semantic similarity and label distributions, we compute the Kullback–Leibler divergence (KLD) between the predicted distributions to quantify their divergence.

Finally, we employ the Jensen–Shannon divergence (JSD) to capture finer-grained distinctions potentially overlooked by the KLD in few‑shot scenarios. Moreover, the normalized JSD $\in [0,1]$ can be directly integrated with semantic similarity scores, obviating the need for KLD normalization.

\section{TopK with Label Distribution Divergence for In-Context Demonstration Selection}

In this section, we begin by presenting the standard TopK retrieval in initial demonstration selection, then re-rank the selected in-context demonstrations by calculating the label distribution divergence with KLD and JSD, to align label distribution with semantic similarity.

\subsection{Semantic Retrieval stage}

This stage initially employs a semantic retriever to calculate cosine similarity $Sim^{i,j}$ between the $i$th test input $T_{test}^{i}$ and the training input $T_{train}^{j}$ where $j$ denotes the $j$th training input in the training set, based on their embeddings $E_{test}^i$ and $E_{train}^{j}$:

\begin{equation}
        Sim^{i,j} = \frac{E_{test}^i \cdot E_{train}^j}{||E_{test}^i|| \; || E_{train}^j||}, 
\label{eq:1}
\end{equation}
followed by ranking the TopK candidates to form a demonstration pool $D_{pool}^i$ for $i$th test input and obtaining the TopK similarity scores $S_{text}^{i,k}$, where $k\in[1,K]$.

\subsection{Label Divergence Estimation stage}

This stage aims to address the potential misalignment between semantic similarity and label distribution by introducing a probabilistic distribution alignment mechanism. Specifically, an SLM is fine-tuned on each $T_{train}^{j}$ initially, and then predicts label distributions for each test input $P_{test}^i$ and each demonstration $P_{pool}^{i, k} \in P_{pool}^i$.

The discrepancy between these distributions is then quantified using KLD:


\begin{equation}
D_{KL}(P_{test}^i || M) = \sum_{c}{P_{test}^i(c) \log\frac{P_{test}^i(c)}{M(c)}},
\end{equation}
\begin{equation}
D_{KL}(P_{pool}^{i, k} || M) = \sum_{c}{P_{pool}^{i,k}(c) \log\frac{P_{pool}^{i,k}(c)}{M(c)}},
\end{equation}

where $c$ denotes the class index and $M=\frac{P_{test}^i + P_{pool}^{i,k}}{2}$ 
represents the midpoint distribution, introduced to mitigate the directional bias inherent in the asymmetric KLD. To ensure the divergence metric balances both the test input and the demonstration pool, we apply JSD to smooth the label distribution differences. This mitigates the impact of potential local outliers while preserving semantically relevant demonstrations, denoted as:

\begin{equation}
\begin{split}
D_{JS}(P_{test}^i \| P_{pool}^{i,k}) = \tfrac{1}{2} \bigl[ 
  D_{KL}(P_{test}^i \| M) \\
  + D_{KL}(P_{pool}^{i,k} \| M) 
\bigr].
\end{split}
\end{equation}

We define the label distribution matching score $S_{label}^{i,k}=1-D_{JS}$, where $D_{JS}=0$ indicates perfect alignment, bounded within [0,1] that can be directly integrated with similarity scores $S_{text}^{i,k}$.

In order to control a trade-off between semantic and distributional alignment, the final ranking employs a hybrid scoring $S_{hybrid}^{i,k}$ strategy, denoted as:
\begin{equation}
\label{eq:5}
S_{hybrid}^{i,k} = \alpha \cdot S_{text}^{i,k} + (1-\alpha) \cdot S_{label}^{i,k},
\end{equation}
where $\alpha \in [0,1]$ is a tunable weight balancing semantic relevance and label distributional consistency.

\section{Experimental Setup}

\textbf{Datasets.} We conduct a comprehensive evaluation across seven text classiffcation tasks for evaluating the generalizability of the proposed TopK+L2D, including \textbf{Binary Classification}: SST-2 \cite{socher2013recursive}, CR \cite{ding2008holistic}, and Subj \cite{wang2018glue}; \textbf{Multi-class Classification}: SST-5 \cite{socher2013recursive} and AgNews \cite{Zhang2015CharacterlevelCN}); and \textbf{Natural Language Inference}: MNLI \cite{williams2018broad} and QNLI \cite{wang2018glue}. Detailed statistics for each dataset are provided in Table~\ref{tab:1}.

\begin{table}[!htbp]
    \centering
    \scriptsize
    \begin{tabular}{ l 
                    *{4}{@{\hspace{4pt}}c@{\hspace{-2pt}}} 
                    c} 
        \hline
        \textbf{Dataset} & \textbf{Train} & \textbf{Val} & \textbf{Test} & \textbf{Labels} & \textbf{Task} \\
        \hline
        \textbf{SST-2} & 6,920 & 872 & 1,821 & 2  & Sentiment Classification \\ 
        \textbf{SST-5} & 8,544 &  1,101 & 2,210 & 5  & Sentiment Classification \\ 
        \textbf{CR} &  3,394 & 0  & 376 & 2  & Sentiment Classification \\ 
        \textbf{Subj} & 8,000 & 0 & 2,000 & 2  & Subjectivity Analysis\\
        \textbf{AgNews} & 120,000 & 0  & 7600 & 4  & Topic Classification \\
        \textbf{MNLI} & 392,702 & 19,647 & 19,643 & 3 & Natural Language Inference \\
        \textbf{QNLI} & 104,743 & 5,463 & 5,463 & 2 & Natural Language Inference \\
        \hline
    \end{tabular}
    \caption{The statistics of datasets. }
    \label{tab:1}
\end{table}

\begin{table*}[!htbp]
\centering
\small
\begin{threeparttable}
\begin{tabular}{lccccccc|c}
\toprule
\textbf{Method} &\textbf{AgNews} & \textbf{CR} & \textbf{SST-2} & \textbf{SST-5} & \textbf{Subj} &\textbf{MNLI} & \textbf{QNLI} & \textbf{Average}  \\
\midrule
\textbf{Random} & 68.12  & 92.82  & 95.22  & 44.93  & 73.95  & 76.41  & 82.87 & 76.33  \\
\textbf{BM25} & 75.77 & 93.62  & 95.00 & 46.38 & 90.85 & 79.08 & 82.67 & 80.48 \\
\textbf{TopK} & 75.71 & 93.62 & \underline{96.05} & \underline{50.23} & \underline{92.45} & \underline{80.01} & 82.76 & 81.55 \\
\textbf{TopK + MDL} & 77.83 & \underline{94.41} & \underline{96.05} & 50.05 & 92.35 & 79.90 & 82.76 & \underline{81.91} \\
\textbf{TopK + ConE} & \textbf{80.95} & 93.88 & 95.61 & 48.91 & 90.75 & 78.61 & \underline{84.26} & 81.85 \\
\midrule
\textbf{TopK + L2D(Ours)} & \underline{78.20} & \textbf{94.68} & \textbf{96.49} & \textbf{54.30} & \textbf{95.15} & \textbf{83.48} & \textbf{85.45} & \textbf{83.96} \\

\bottomrule
\end{tabular}
\end{threeparttable}
\caption{Performance comparison between L2D and baseline models in accuracy(\%) across seven tasks on Qwen2.5-7B-Instruct. \textbf{Bold} numbers indicate the best performance, while \underline{underline} values denote the second-best results.}
\label{tab:comparison}
\end{table*}

\textbf{Large language models.} We evaluate our method across a range of LLMs, employing \texttt{Qwen2.5-7B-Instruct} \cite{qwen2.5} as the primary model for most experiments. To assess the scalability and generalizability of the L2D
method, we further conduct experiments using LLMs of varying sizes, from 2B to 14B parameters. These include \texttt{Gemma2-2B-it} \cite{gemma_2024}, \texttt{Phi3-mini-128k-Instruct} \cite{abdin2024phi}, \texttt{LLaMA3-8B-Instruct} \cite{llama3modelcard}, \texttt{Gemma2-9B-it} \cite{gemma_2024}, and \texttt{Qwen2.5-14B-Instruct} \cite{qwen2.5}. 

\textbf{Small language models.} For label distribution estimation, we adopt \texttt{RoBERTa-base} \cite{DBLP:journals/corr/abs-1907-11692} as the primary model. To assess the influence of SLMs on the performance of L2D, we additionally incorporate two classic SLMs: \texttt{BERT-base-uncased} \cite{devlin2019bert} and \texttt{DeBERTa-v3-base} \cite{he2021deberta}.

\textbf{Baselines.} We primarily compare our method against five widely adopted baselines for in-context demonstration selection.

\begin{itemize}
    \item \textbf{Random} denotes that the in-context demonstrations are randomly selected.

    \item \textbf{BM25} \cite{robertson2009probabilistic} computes word-overlap similarity between training samples and the test input, selecting the most similar samples as demonstrations.

    \item \textbf{TopK} \cite{liu2022makes} selects the nearest neighbors from training samples for a given test input as its corresponding in-context demonstrations.

    \item \textbf{TopK + MDL} \cite{wu2023self} adopt a select-then-rank framework, in which demonstrations retrieved via the TopK method are ranked according to the Minimum Description Length (MDL) principle.

    \item \textbf{TopK + ConE} \cite{peng2024revisiting} is a data- and model-dependent demonstration selection method, which posits that effective demonstrations are those that minimize the conditional entropy of the test input under the inference model.
\end{itemize}


\textbf{Evaluation Metrics.} In our experiments, we use the accuracy rate (ACC) for evaluation.

\textbf{Experimental Details.} To ensure a fair comparison, we follow the experimental setup introduced by \citet{peng2024revisiting}. All experiments are conducted on two RTX 4090 GPUs, with a fixed random seed to ensure reproducibility. Specifically, we use the TopK method to retrieve 30 candidate demonstrations for each test sample, which are then re-ranked using our proposed L2D method. Prompt templates are adopted from \citet{lu2022fantastically} and \citet{wu2023self}. We introduce a tunable weight $\alpha$ to balance the semantic similarity score and the label distribution consistency score. Based on empirical validation, $\alpha$ is fixed at 0.5, the detail is illustrated in Figure \ref{fig:alpha}. The main comparative experiments are conducted under an 8-shot ICL setting, using \texttt{gte-base-en-v1.5} \cite{li2023gte} as the default model for semantic retrieval. For label distribution estimation, we use RoBERTa-base as the default SLM. The training set is split in an 8:2 ratio for training and validation. Fine-tuning is performed using the Huggingface Trainer API to evaluate SLM performance. 


\section{Main Results}

We evaluate the effectiveness of the proposed TopK + L2D on seven classification tasks. The experimental results demonstrate that:

\textbf{Our method consistently outperforms other baselines on almost all tasks.} The proposed TopK + L2D outperforms the existing SOTA select-rerank frameworks as shown in Table \ref{tab:comparison}, including TopK+ConE and TopK+MDL, across six out of seven tasks, yielding an average accuracy improvement of 2.11\% and 2.05\%, respectively. In addition, we also conduct a paired t-test across 10 seeds comparing TopK+L2D, TopK+ConE and TopK+MDL. The results indicate that our method (TopK+L2D) significantly outperformed TopK+ConE (t$=$11.6815, p$<$0.01) and TopK+MDL (t$=$7.3556, p$<$0.01). Compared to the standard TopK method, this improvement increases to 2.41\%, highlighting the effectiveness of incorporating label distribution divergence in enhancing the classification performance of LLMs. For NLI tasks, our method achieves substantial improvements, 3.47\% on MNLI and 1.19\% on QNLI, compared to the second-best methods, demonstrating its superior performance on semantically challenging tasks.

\begin{figure}[!htbp]
  \centering
  \includegraphics[width=\columnwidth]{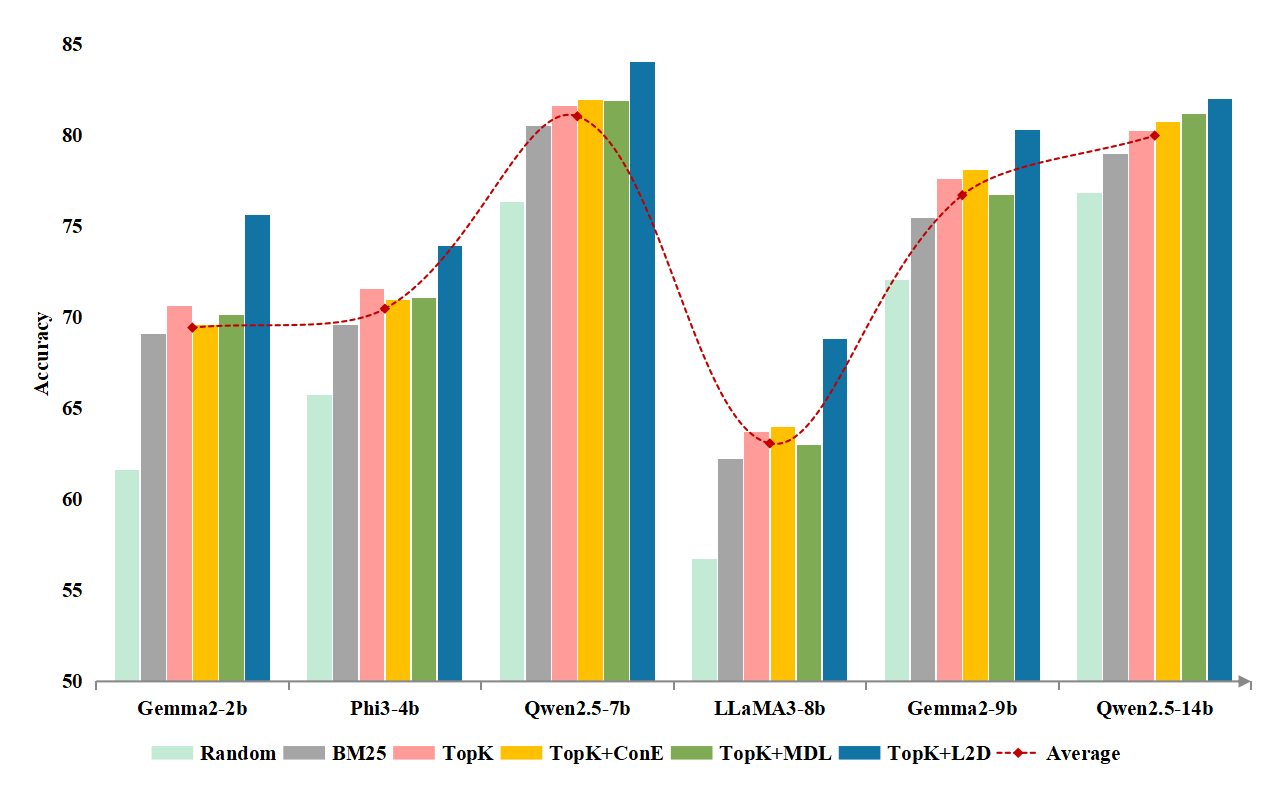}
  \caption{The average accuracy of LLMs on seven tasks at different scales shows that our method consistently improves performance across various models. The red dotted line illustrates the averaged performance of all methods across different model scales.}
  \label{fig:model_comp}
\end{figure}

\begin{figure*}[!t]
     \centering
     \begin{subfigure}[b]{0.47\textwidth}
         \centering
         \includegraphics[width=\textwidth]{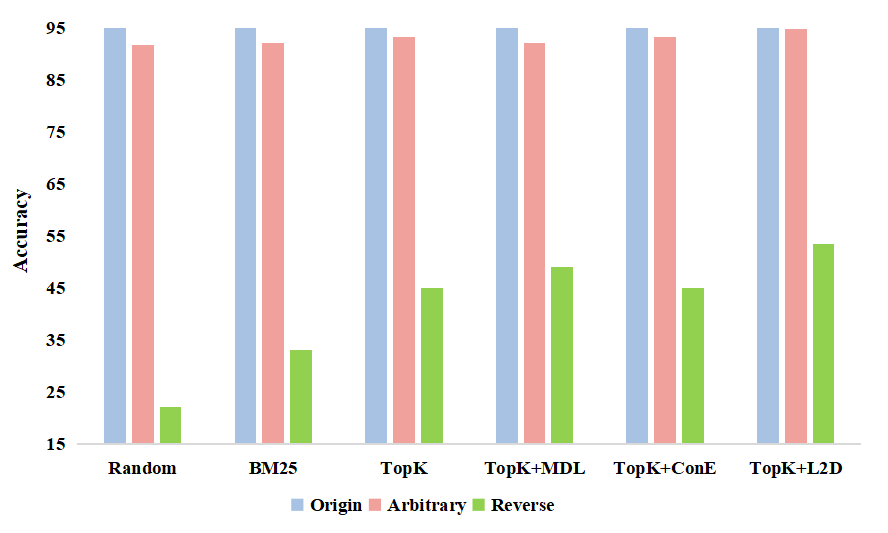}
         \caption{SST-2}
         \label{fig:sst-ar}
     \end{subfigure}
     \begin{subfigure}[b]{0.47\textwidth}
         \centering
         \includegraphics[width=\textwidth]{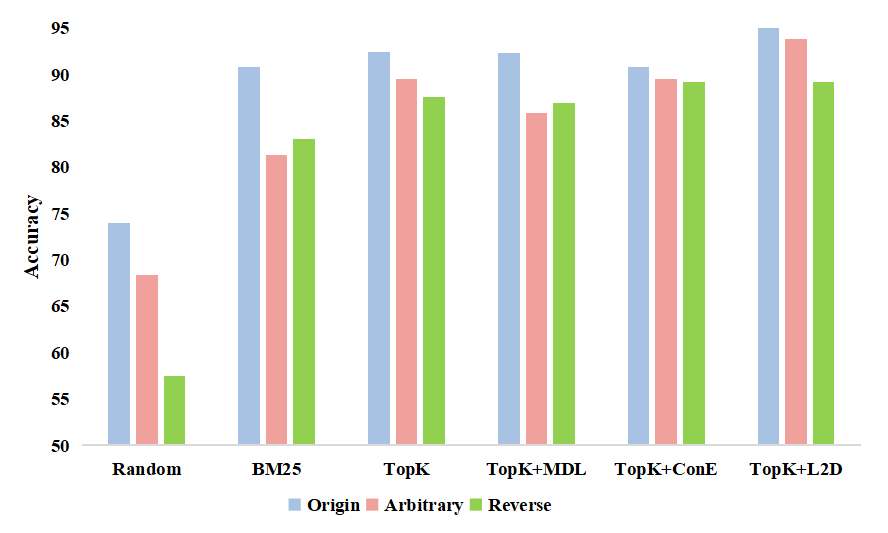}
         \caption{Subj}
         \label{fig:subj-ar}
     \end{subfigure}
     
     \begin{subfigure}[b]{0.47\textwidth}
         \centering
         \includegraphics[width=\textwidth]{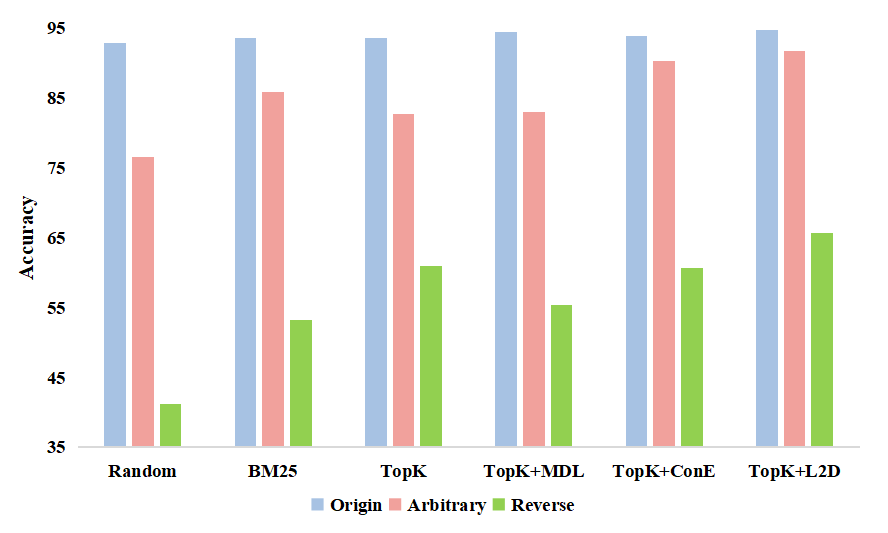}
         \caption{CR}
         \label{fig:cr-ar}
     \end{subfigure}
     \begin{subfigure}[b]{0.47\textwidth}
         \centering
         \includegraphics[width=\textwidth]{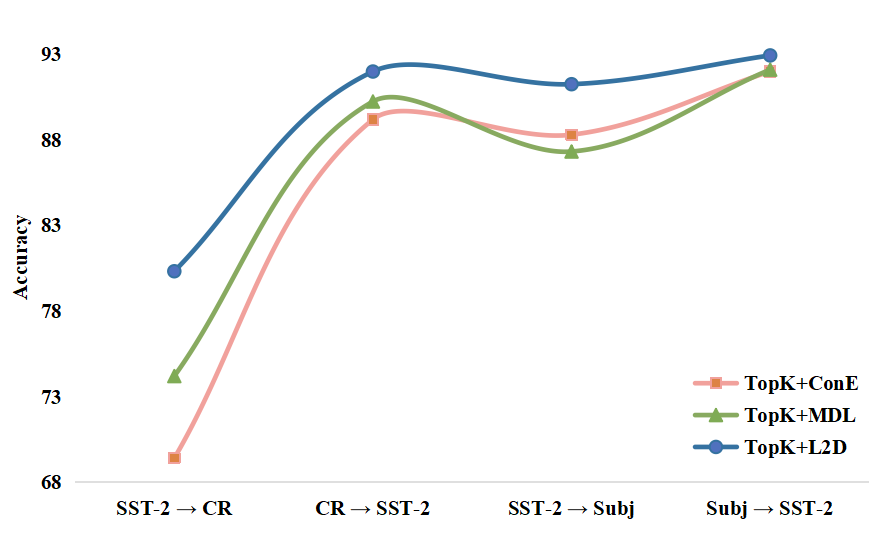}
         \caption{Out-of-domain}
         \label{fig:ood}
     \end{subfigure}
     
        \caption{ 
        (a) - (c) Performance comparison between original, arbitrary and reverse labels in in-context demonstrations across SST-2, Subj and CR. (d) Performance of our method on out-of-domain demonstration pools. `A$\rightarrow$B' indicates that the demonstration pool is sourced from dataset `A' while evaluation is conducted on dataset `B'.}
        \label{fig:analysis-1}
\end{figure*}

\textbf{Our method brings consistent performance improvements across various models and model scales.} We evaluate various demonstration selection methods across multiple LLMs (i.e., Gemma2, Phi3, Qwen2.5, and LLaMA3) and model scales ranging from 2B to 14B, as illustrated in Figure~\ref{fig:model_comp}. Experimental results demonstrate that our method consistently outperforms baseline approaches across various models and scales, most notably with Gemma2-2B, where it achieves average accuracy gains of 5.53\% and 6.08\% over TopK + MDL and TopK + ConE, respectively. It is also noteworthy that LLM performance does not always scale positively with model size, as shown by the red dotted line in Figure \ref{fig:model_comp}. This observation contrasts with the findings reported in \citet{peng2024revisiting}, but aligns with the conclusions drawn in \citet{wang2023label}.

\section{Analysis}

We conduct extensive analyses with Qwen2.5-7B-Instruct on classification tasks to further investigate the effectiveness and generalizability of our method. 

\textbf{Our method is robust to arbitrary and reversed labels.} Prior studies by \citet{min2022rethinking} and \citet{yoo2022ground} suggest that the correctness of text-label pairs in in-context demonstrations has limited influence on LLM performance. To further examine the robustness of our method under noisy label settings, we conduct experiments in which the correct labels are either replaced with arbitrary symbols (e.g., `foo' and `bar') or reversed (e.g., flipping a `positive' label to `negative'). Figures \ref{fig:analysis-1}(a)-(c)
illustrate that TopK+L2D consistently outperforms other baseline methods under both conditions. These results suggest that label correctness in in-context demonstrations can substantially impact LLM performance, contrary to earlier assumptions. Interestingly, most methods are less affected by label reversal in the Subj dataset compared to SST-2 and CR. This may be attributed to the inherently more nuanced distinction between subjective and objective expressions, whereas reversing positive and negative sentiment labels directly conflicts with the semantic intent of the input-label pairs.

\textbf{Our method works for out-of-domain (OOD) demonstration pools.} While previous results have demonstrated the effectiveness of our method in in-domain demonstration pools, we now evaluate its generalizability in OOD settings, where demonstrations are drawn from domains different from the target task. Specifically, we conduct experiments across three domains, Movie Review (SST-2), Customer Review (CR), and Subjectivity Analysis (Subj), by constructing demonstration pools from one dataset and evaluating performance on the others. This setup enables us to investigate the robustness of our approach when faced with domain shifts. As shown in Figure \ref{fig:analysis-1}(d), our method consistently outperforms other select-rerank frameworks across all OOD scenarios, achieving average performance gains of 3.16\% and 4.4\% over TopK + MDL and TopK + ConE, respectively.

\begin{figure*}[!t]
     \centering
     \begin{subfigure}[b]{0.47\textwidth}
         \centering
         \includegraphics[width=\textwidth]{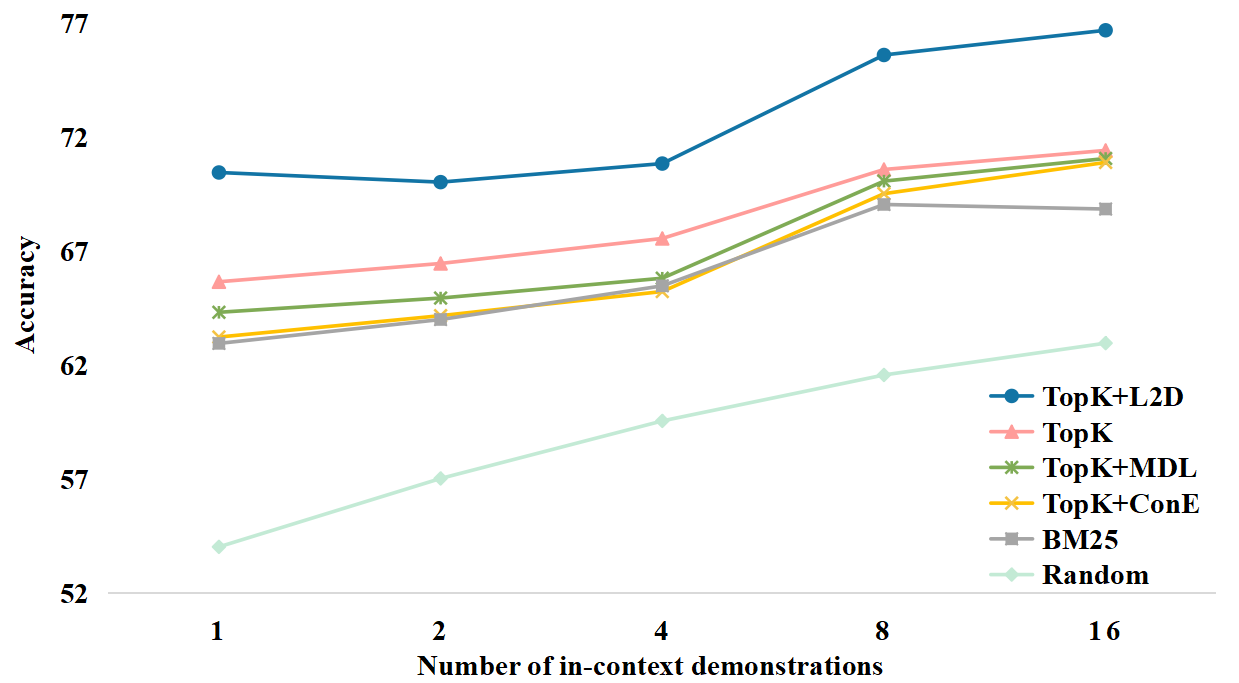}
         \caption{}
         \label{fig:n-shot}
     \end{subfigure}
     \begin{subfigure}[b]{0.47\textwidth}
         \centering
         \includegraphics[width=\textwidth]{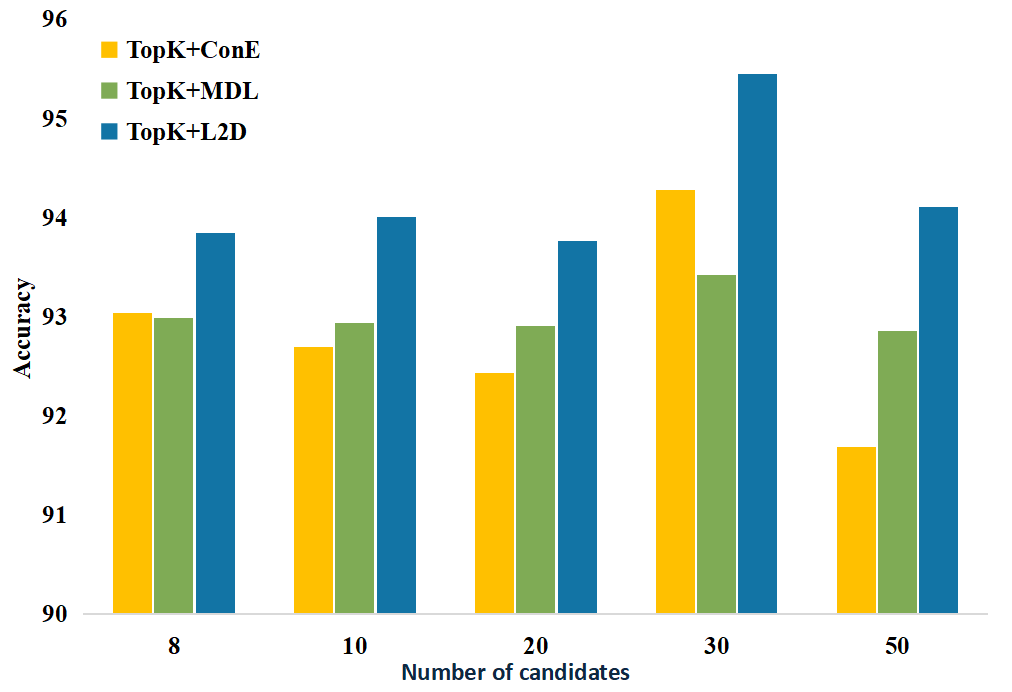}
         \caption{}
         \label{fig:candidate}
     \end{subfigure}
     
        \caption{Performance comparison between different number of (a) in-context demonstrations and (b) candidates in semantic retrieval stage. }
        \label{fig:analysis-2}
\end{figure*}

\section{Impact of hyperparameters} \label{sec:impact of hyperparameters}

In this section, we conduct a comprehensive analysis of how different hyperparameter settings influence the performance of the TopK+L2D framework with different LLMs.

\begin{table*}[!t]
    \centering
    \small

    \begin{tabular}{lcccc|l} 
        \hline
        \textbf{Method} & \textbf{CR} & \textbf{SST-2} & \textbf{SST-5} & \textbf{Subj} & \textbf{Average} \\
        \hline
        \textbf{L2D-BERT-Gemma2} & 92.55 & 94.51 & 34.25 & 94.90 & 79.05(+\textbf{1.23}) \\
        \textbf{L2D-BERT-Phi3} & 89.89 & 94.34 & 42.03 & 92.04 & 79.58(+\underline{0.61}) \\
        \textbf{BERT} & 91.18 & 91.62 & 54.27 & 97.62 & 83.67($\uparrow$2.98) \\
        \hline
        \textbf{L2D-RoBERTa-Gemma2} & 92.55 & 94.18 & 35.38 & 95.20 & 79.33(+\textbf{0.95}) \\
        \textbf{L2D-RoBERTa-Phi3} & 90.69 & 94.67 & 41.89 & 91.79 & 79.76(+\underline{0.43}) \\
        \textbf{RoBERTa} & 92.35 & 91.76 & 55.09 & 96.88 & 84.02($\uparrow$2.63) \\
        \hline
        \textbf{L2D-DeBERTa-Gemma2} & 93.09 & 95.17 & 37.24 & 95.60 &  80.28(+\textbf{0}) \\
        \textbf{L2D-DeBERTa-Phi3} & 90.96 & 94.56 & 42.84 & 92.39 & 80.19(+\underline{0}) \\ 
        \textbf{DeBERTa} & 93.24 & 95.38 & 60.23 & 97.75 & 86.65($\uparrow$0) \\
        \hline
    \end{tabular}
    \caption{Performance comparison among different small language models with L2D on LLMs.  (+) indicates the relative improvements achieved when combining our L2D method with various small language models on Gemma2-2B-it  and Phi3-mini-128k-Instruct (scores in \textbf{bold} and in \underline{underline}, respectively). ($\uparrow$) denotes the relative performance differences observed among the standalone small language models.}
    \label{tab:slm}
\end{table*}

\textbf{Impact of the number of in-context demonstrations.}
We begin by examining how the number of in-context demonstrations influences model performance. Specifically, we incrementally increase the number of demonstrations from 1 to 16 and evaluate the results using Gemma2-2B-it. The average accuracy of all methods is calculated and presented in Figure \ref{fig:analysis-2}(a).

  

We observe that \textbf{increasing the number of in-context demonstrations consistently leads to improved performance on average, suggesting a positive correlation between LLM performance and the number of the provided demonstrations}. Notably, our method consistently outperforms all baseline approaches across various settings. Moreover, as the number of in-context demonstrations increases, the performance gains of our method also grow, achieving significant improvements of 5.48\% and 6.11\% over TopK+MDL and TopK+ConE, respectively. These results highlight the robustness and scalability of our approach with extended context.

\textbf{Impact of the number of TopK candidates.} We examine how varying the number of candidate demonstrations retrieved from the demonstration pool via the TopK method during the semantic retrieval stage affects performance, using Qwen2.5-7B-Instruct as the evaluation model. Specifically, we vary the number of candidate demonstrations retrieved during the semantic retrieval stage from 8 (i.e., equal to the number of in-context demonstrations used) up to 50 across SST-2, Subj and CR datasets, to explore how a larger candidate pool influences the overall performance, as shown in Figure \ref{fig:analysis-2}(b).

We observe that increasing the number of TopK candidates does not lead to performance improvements until the candidate size reaches 30. However, performance degrades when the candidate size increases to 50. Moreover, a larger candidate set introduces additional latency during the semantic retrieval stage. We hypothesize that \textbf{overly large candidate pools introduce more semantically distant ``noise'' that even a high quality L2D reranker cannot fully eliminate, thereby slightly hurting accuracy and increasing latency}. Therefore, we set the default number of TopK candidates to 30, balancing performance gains and computational efficiency. Notably, our method consistently outperforms other baselines across all candidate settings, demonstrating its robustness and effectiveness with extended demonstration candidates.

\textbf{Impact of small language models.} Given that label distribution plays a critical role in the reranking stage, we hypothesize that \textbf{higher accuracy achieved by the SLM in generating label distributions can enhance the overall performance of LLMs}. To validate this assumption, we analyze how the performance of the proposed L2D method correlates with the standalone accuracy of different SLMs, namely BERT-base-uncased, RoBERTa-base, and DeBERTa-v3-base. We choose two LLMs, Gemma2-2B-it and Phi3-mini-128k-Instruct, from distinct families, featuring diverse architectures and instruction‑tuning protocols, and evaluate them on four tasks (CR, SST-2, SST-5, and Subj) to maintain tractable computational requirements.

As shown in Table~\ref{tab:slm}, the performance of TopK+L2D improves consistently as the accuracy of the SLM increases, supporting our hypothesis of a positive correlation between SLM predictive accuracy and LLM performance in ICL. Specifically, DeBERTa-v3-base achieves the highest standalone accuracy among the evaluated SLMs, outperforming RoBERTa and BERT by 2.63\% and 2.96\%, respectively. This trend extends to our method: L2D-DeBERTa-Phi3 and L2D-DeBERTa-Gemma2 surpass their RoBERTa-based counterparts by 0.43\% and 0.95\%, and their BERT-based counterparts by 0.61\% and 1.23\%, respectively. Interestingly, we also observe that SLMs outperform LLMs on average across most tasks in standalone classification settings, indicating that there is still considerable room for improvement in ICL methods.

\begin{figure}[!htbp]
  
  \includegraphics[width=0.47\textwidth]{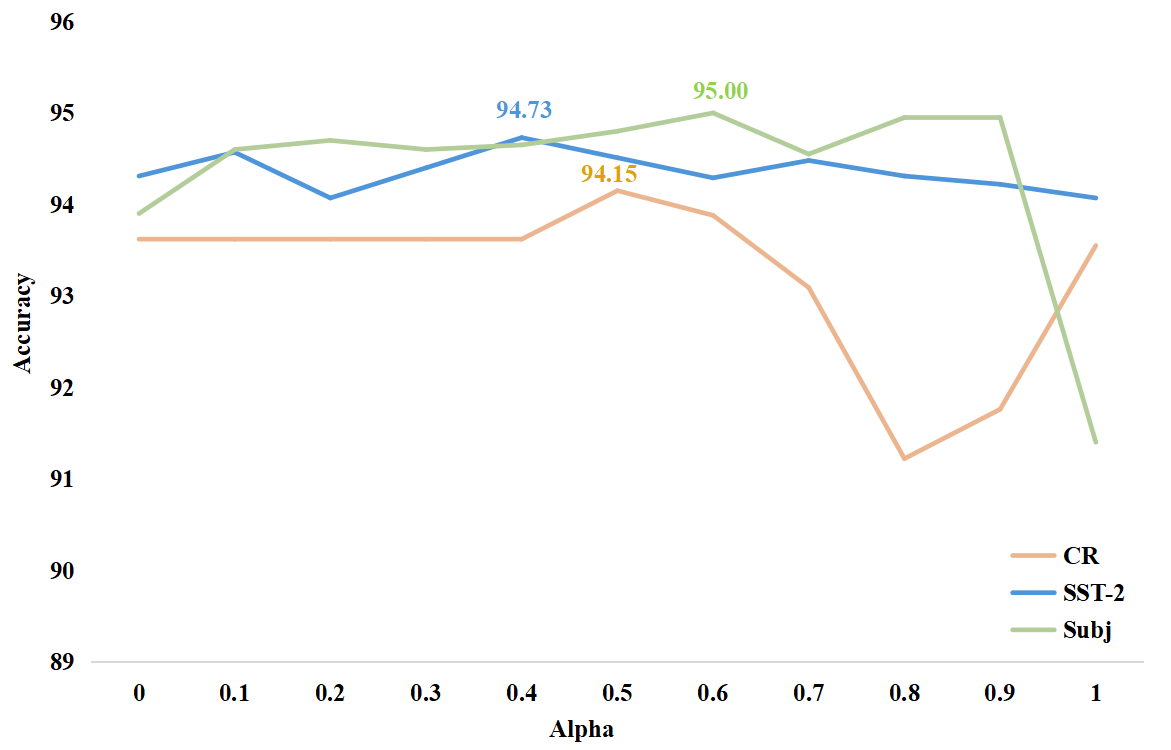}
  \caption{Performance comparison of different $\alpha$ settings.}
  \label{fig:alpha}
\end{figure}

\textbf{Impact of tunable weight $\alpha$ and ablation studies.} We introduce a tunable weight $\alpha$ to control the balance between the semantic similarity score ($S_{\text{text}}^{i,k}$) and the label distribution matching score ($S_{\text{label}}^{i,k}$) in Eq~\ref{eq:5}. To investigate the influence of this trade-off, we vary $\alpha$ from 0 to 1 in increments of 0.1 with Qwen2.5-7B-Instruct, thereby exploring the relationship between semantic similarity score and label distribution matching score in the overall scoring function, as shown in Figure \ref{fig:alpha}.

We observe that LLMs generally achieve better accuracy when the value of $\alpha$ is set within the range of 0.4 to 0.6, indicating that \textbf{optimal performance can be obtained by appropriately balancing semantic similarity and label distributional consistency}. Based on this observation, we set the default value of $\alpha$ to 0.5.

In addition, we evaluate two extreme cases by setting $\alpha$ to 1 (i.e., relying solely on semantic similarity) and 0 (i.e., relying solely on label distributional consistency) to conduct an ablation study of the proposed TopK+L2D method. Notably, when $\alpha$ is set to 0, the candidates are still retrieved using the TopK method but are no longer sorted by their semantic similarity scores, this setting is denoted as \textbf{w/o Sem} in Table~\ref{tab:abl}.

\begin{table}[htb]
    \centering
    
    \small
    \begin{tabular}{lccc} 
        \hline
        \textbf{Method} & \textbf{CR} & \textbf{SST-2} & \textbf{Subj}  \\
        \hline
        \textbf{TopK+L2D} & 94.68 & 96.49 & 95.15  \\
        \hline
        \textbf{w/o Sem} & 93.62($\downarrow$1.06) & 94.31($\downarrow$2.18) & 93.90($\downarrow$1.25)  \\
        \textbf{w/o L2D} & 93.55($\downarrow$1.13) & 94.07($\downarrow$2.42) & 91.40($\downarrow$3.75) \\
        \hline
    \end{tabular}
    \caption{Ablation study on three datasets. \textbf{w/o Sem} refers to a setting in which demonstrations are ranked solely based on
label distributional consistency, omitting semantic similarity scoring. \textbf{w/o L2D} denotes the vanilla top-K selection method
without the application of L2D-based reranking.}
    \label{tab:abl}
\end{table}

We observe that the performance of TopK+L2D degrades proportionally when each component is removed. Specifically, the average performance drops by 1.50\% when semantic sorting is excluded, and by 2.30\% when the label distributional consistency score is removed. These results highlight that both components are essential and jointly contribute to the optimal performance of our method.

\section{Conclusion}

In this paper, we propose a two-stage demonstration selection method, TopK+L2D, specifically designed to enhance the performance of LLMs on text classification tasks. Our approach addresses the challenge that labels are often not entirely independent of instances, particularly in scenarios with noisy data or label ambiguities. By jointly considering semantic similarity and label distribution alignment, our method selects more effective in-context demonstrations. 



\section*{Limitations and future work}
We acknowledge several limitations in our work. (1) The selection of LLMs is constrained to model scales between 2B and 14B due to limited computational resources. (2) While our study primarily investigates the impact of label distribution divergence in in-context learning, part of the strong performance can also be attributed to the underlying TopK retrieval mechanism, as highlighted in the ablation study (Table~\ref{tab:abl}). A more thorough evaluation of the retrieval component is left for future work. (3) Our experiments focus mainly on text classification tasks. Future work will explore the applicability of our approach to other tasks, such as text generation, to further examine its generalizability.

\bibliography{aaai2026}



\end{document}